\documentclass{article}

\usepackage{times}
\usepackage{graphicx} 
\usepackage{subfigure} 

\usepackage{natbib}

\usepackage{algorithm}
\usepackage{algorithmic}

\usepackage{hyperref}

\usepackage[accepted]{icml2017}

\usepackage{graphicx}
\graphicspath{ {images/} }

\icmltitlerunning{Probabilistic and Regularized Graph Convolutional Networks}

\begin{document} 

\twocolumn[
\icmltitle{Probabilistic and Regularized Graph Convolutional Networks (L42 Michaelmas 2017)}




\begin{icmlauthorlist}
\icmlauthor{Sean Billings}{cam}
\end{icmlauthorlist}

\icmlaffiliation{cam}{University of Cambridge, Cambridge, United Kingdom}

\icmlcorrespondingauthor{Sean Billings}{sb2219@cam.ac.uk}

\icmlkeywords{boring formatting information, machine learning, ICML}

\vskip 0.3in
]



\printAffiliationsAndNotice{}  

\begin{abstract} 

This paper explores the recently proposed \textit{Graph Convolutional Network} architecture proposed in \cite{tkipf} The key points of their work is summarized and their results are reproduced. Graph regularization and alternative graph convolution approaches are explored. I find that explicit graph regularization was correctly rejected by \cite{tkipf}. I attempt to improve the performance of GCN by approximating a k-step transition matrix in place of the normalized graph laplacian, but I fail to find positive results. Nonetheless, the performance of several configurations of this GCN variation is shown for the Cora, Citeseer, and Pubmed datasets.\\

\end{abstract} 

\section{Introduction}

Data structures such as graphs serve as an interesting exploratory realm for atypical neural network applications. Because of the success of deep learning strategies on traditional supervised learning domains such as image recognition and machine translation, it is thought that novel neural network architectures can find similarly revolutionary results for other data structures such as graphs. \\

In \cite{tkipf} a novel neural architecture is revealed to perform very well at classifying nodes in graph data structures. This network is shown to have state-of-the-art performance for several citation datasets including citeseer and pubmed as well as several other standard datasets. The key insight of this paper is the design of a graph convolutional network (GCN) layer that allows for the implicit learning of network structure, while simultaneously incorporating node features. \\

This paper will review the formal background for the network in \cite{tkipf}, as well as develop and evaluate several new techniques inspired from the GCN and clustering literature. It is thought by the author that more adaptive graph convolutional network layers could potentially improve the performance of the vanilla GCN implementation. \\

The paper will proceed as follows; Section 2 of this paper will cover related work and link to several key ideas that relate to GCN. Section 3 of this paper will outline the theory that formalizes the GCN and GCN-variants proposed in this paper. Section 4 is composed of experiments that reproduce the GCN results and explore the GCN-variants of this paper. Section 5 will conclude this paper, and account for the shortcomings of these GCN-variants. \\

\section{Related Work} 

Applying neural networks to graph structured data has been explored by many authors over the past decade. Here, I highlight some of the key inspirations of, and related work to, the GCN architecture. \\

\cite{Belkin2006} presents the underlying graph regularization that ends up being rejected by \cite{tkipf}. They outline what is termed Manifold Regularization as an additive loss term that helps neural networks to discover graph structure. This form of explicit regularization was shown to have very nice theoretical properties and was applied as a regularization tool for several machine learning techniquess such as regression and Support Vector Machines .
 
\cite{Bruna2014} look at applying Convolutional Neural Network architectures to graph-structured classification problems. They do so by considering the information passing through a layer of a network as a signal to be transformed with graph spatial information. They perform a spatial re-construction of the signal by incorporating a diagonalization of the eigenvalues of the associated graph laplacian.   \\

\cite{Defferrard2016} also attempt to extend CNN to Graphs. Similarly to \cite{tkipf}, they look at how Chebyshev polynomials can be used to quickly approximate and encode graph structure. Using these Chebyshev polynomials, the authors define graph transformations on the dataset that act in such a way to transform an input vector x according to the graph structure. These authors focus on coarsening and pooling these graph signals in such a way that the network can more effectively learn and encode the dataset. \\

\cite{tkipf} have presented a very powerful graph neural architecture that is the current state-of-the-art. However, I believe this is the beginning rather than the end of exploring these kinds of GCN architectures.

\section{Graph Convolutional Networks}

In this section, I present the theory underlying GCN and our proposed variants in order to set the foundations for our experiments.

\subsection{Graph Data Structure}

I review several Mathematical Graph objects to formalize the GCN approach. Denote a graph $ G = (V,E) $ as a set of vertices $V$ and edges $E$. Associated with $G$ is the adjacency matrix $A$, where $A_{ij} > 0 $ iff $(i,j) \in E$ and a degree matrix $D$ with $D_{ii} = \sum_j A_{ij}$ .  Extending this, a self-connection adjusted adjacency matrix  $ \tilde{A} = A + I $ can also be used to define the adjusted degree matrix $\tilde{D} $ where $ \tilde{D}_{ii} = \sum_j $ \~{A}$_{ij} $. The graph Laplacian is defined as $\Delta = D - A $. The normalized graph laplacian can then be formulated as $\tilde{D}^{-1/2} \tilde{A} \tilde{D}^{-1/2}$.\\

\subsection{Graph Convolutional Layers}

Using Chebyshev polynomials to approximate the spectral convolutions of the graph laplacian, the authors of \cite{tkipf}, \cite{Defferrard2016} show that a linear graph convolution acting on a given node can be approximated with 

\begin{equation}
g_{\theta} \star x \sim \theta(\tilde{D}^{-1/2} \tilde{A} \tilde{D}^{-1/2}) x
\label{eq:graphconv}
\end{equation}

Let us denote the activations of the l'th layer of our neural network as $ H^{(l)} $, and the corresponding weight matrix as $ W^{(l)} $. The activations for a layer of our GCN can be defined as 

\begin{equation}
H^{(l+1)} = \sigma(\tilde{D}^{-1/2} \tilde{A} \tilde{D}^{-1/2} H^{(l)} W^{(l)})
\label{eq:adjacencylayer}
\end{equation}

In the above $\sigma$ is a nonlinear activation function such as ReLu or Tanh, and the operation can be seen as a non-linear generalization of equation \ref{eq:graphconv} . \\

The authors argue that their first order convolution approximation in equation \ref{eq:adjacencylayer} should reduce over-fitting. The authors argue that repeated operation of this convolution operator is equivalent to repeated applications of a transition matrix to the network. If we would like to control the number of steps per operation, one alternative structure that we could look at is to use a generalized probabilistic transition matrix $ P^k $ that approximates the k'th order markov chain for the graph transition matrix. In this way we can (1) normalize our graph structure (2) generalize the framework beyond immediately adjacent nodes. I define this operation as  \\

\begin{equation}
H^{(l+1)} = \sigma(P^k H^{(l)} W^{(l)})
\label{eq:transitionlayer}
\end{equation}

This probabilistic approach to modelling network structures at varying depth is inspired by the work in \cite{Pons2006}. I generate $P^k$ by performing a set of random walks of depth k for each node in the graph. This architecture is referred to as PGCN.

\subsection{Training GCN} 

For the labelled set of nodes $Y_L$, the loss function for training GCN in \cite{tkipf} is the standard cross-entropy loss

\begin{equation}
L_0  = - \sum_{l \in Y_L} Y_l \ln (f(X_l))
\label{eq:labelloss}
\end{equation}

Although not in the original paper, in the implementation of GCN, weight decay regularization on the first layer of the GCN is also added to the loss function. The form of weight decay regularization is 

\begin{equation}
L_{wd} = \frac{1}{2} \lambda \sum_i w_i^2 
\label{eq:weightdecaryloss}
\end{equation}

\cite{tkipf} aims to avoid explicitly enforcing graph constraints. However, one could consider combining the standard loss function $L_{GCN} (Y, f(X) ) $ measuring the prediction error in each vertex class label with a graph specific regularization term $L_{graph}(A,F(X)) $ that penalizes neighbouring vertices with differing labels. The graph regularization term referenced in \cite{tkipf} and proposed in \cite{Belkin2006} is defined as follows

\begin{equation}
L_{graph}  = \sum_{i,j} A_{ij} \| f(X_i) - f(X_j) \|^2 = f(X)^T \Delta f(X)
\label{eq:graphreg}
\end{equation}

This regularization has interesting properties as shown in \cite{Belkin2006}. It is essentially an optimization strategy that tries to enforce the assumption that adjacent nodes should have similar labels. Notably, this regularization could hypothetically enforce a clustering structure even when the graph vertices do not have other meaningful features associated with them. \\

\section{Experiments}

In this section, I outline several experiments to explore some proposed GCN-variant architectures. 

\subsection{Reproduction of Kipf et al. (2017)}

I recreate the results of \cite{tkipf} by running their proposed GCN implementation on a set of standard citation datasets. The aim is to set a performance benchmark. This benchmark represents the current state of the art for graph processing applications. \\

\begin{table}[h]
\caption {Performance of GCN} \label{tab:title}
\label{tab:LDAConvTopics}
\begin{center}
\begin{tabular}{ |c|c|c|c|c|c|c|c|c|c|c|} 
\hline
Models & Citeseer & Cora & pubmed  \\
 \hline
GCN (Kipf) & 71.4 & 82.2 & 79.3  \\
\hline
\end{tabular}
\end{center}
\end{table}

\subsection{Probabilistic GCN }

In this experiment, I experiment with inserting augmented graph representations into the GCN architecture. For convolutional networks, the convolution operation defines how many neighbouring cells are being considered when performing convolutional operations. The k'th order probabilistic transition matrix approach to graph convolutions defined in equation  \ref{eq:transitionlayer} can be seen as the graph analog for defining this neighbourhood. Whereas for 2-dimensional images we consider pixels within the two dimensional range of $m$ x $n$, for graphs we consider the information produced by a graph diffusion process of k time steps. This is equivalent to a markov chain of k'th order, and can be approximated with random walks on a graph as in \cite{Pons2006}. \\

This Probabilistic GCN (PGCN) model, with layers defined as in equation \ref{eq:transitionlayer}, is run on the Citeseer, Cora, and Pubmed datasets in order to explore how the diffusion depth k influences accuracy.

\begin{table}[h]
\caption {PGCN Cross Validation Results}
\label{tab:PGCN}
\begin{center}
\begin{tabular}{ |c|c|c|c|c|c|c|c|c|c|c|} 
\hline
Models & Citeseer & Cora & pubmed  \\
 \hline
PGCN (1) & 70.6 & 80.8 & 78.0  \\
PGCN (2) & 70.3 & 80.4 & 78.4  \\
PGCN (3) & 69.9 & 80.4 &  78.4 \\
PGCN (4) & 70.2 & 79.8 &   78.6 \\
PGCN (5) & 69.7 & 79.7 &  79.0 \\
\hline
\end{tabular}
\end{center}
\end{table}

Unfortunately, the performance of this approach is lacking. Firstly, rounding errors could be a major influencer of performance. Because the k'th order probabilistic matrices are less sparse than the laplacian, more multiplications and additions will be required, and there is a higher margin for error. When the actual graph clusters are compact (very few hops between nodes in a cluster) and the probabilistic approximation is of higher order than the actual clusters, this will likely be the case. In contrast, when the graph cluster structure is deep, as in the Pubmed dataset, it does seem that higher order transition matrices are more effective. The potential for tuning the graph structure is desirable, and the performance of these techniques in the best case does approximate the vanilla GCN accuracy. However, a defining example of superior performance is absent. \\

\subsection{GCN with Graph Clustering Regularization}

The regularization discuessed in \cite{tkipf} focuses on penalizing neighbouring nodes in the graph with heterogenous class labels. This experiment looks at how that graph regularization strategy impacts the performance of the vanilla GCN architecture.  This variant is referred to as Regularized GCN (RGCN). \\

This experiment looks at the accuracy of GCN and RGCN over a large period of training epochs. RGCN requires longer training cycles. In other words, GCN without the graph regularization tends to trigger early stopping conditions far before a Regularized GCN equivalent would be fully trained. The configurations for this variant are a epoch limit of 5000, with early stopping conditions only considered after 30 epoch for GCN and 1500 epoch for RGCN. \\

\begin{table}[h]
\caption {Test accuracy and epoch count of GCN and RGCN} 
\label{tab:GCN}
\begin{center}
\begin{tabular}{ |c|c|c|c|c|c|c|c|c|c|c|} 
\hline
Models& Citeseer & Cora & pubmed  \\
 \hline
GCN & 71.4 (299e) & 82.2 (359e) & 79.3  (197e)\\
\hline
RGCN  & 71.5 (2140e) & 81.7 (2850e) & 78.6 (1555e) \\
\hline
\end{tabular}
\end{center}
\end{table}

The goal of regularization is to reduce overfitting in the data, and judging by the early stopping of the original GCN model, it does seem to fit the data quickly with potential for overfitting. However, this experiment serves to support the argument made by \cite{tkipf} that explicit graph structure regularization may not be necessary. Here, there does not seem to be any meaningful accuracy improvements. This approach does seem to allow the model to train for longer without triggering early stopping conditions. This leads me to believe that there is potentially some form of regularization that could help improve performance by reducing the tendency to overfit of vanilla GCN. However, more experimentation is needed to figure out what form of regularization that might be. \\ 

\subsection{GCN with Modularity Regularization}

Modularity optimization is a standard technique used in several clustering frameworks such as in \cite{Pons2006}, \cite{Blondel2008}. Although potentially even more computational expensive than the above technique as a regularization term, it may be more effective. This author leaves modularity regularization as potential future work.

\subsection{PGCN with Graph Clustering Regularization}

I propose a regularized form of PGCN that attempts to incorporate graph clustering regularization. This regularization replaces the graph laplacian in \ref{eq:graphreg} with the probabilistic transition matrix as a k'th order version of cluster regularization. The updated regularization term is 

\begin{equation}
L_{PRGCN}  =  L_0 + L_{wd} + f(X)^T P^k f(X)
\label{eq:graphreg}
\end{equation}

I evaluate the cross-validated performance PRGCN to parallel the evaluation in table \ref{tab:PGCN}.

\begin{table}[h]
\caption {PRGCN Cross Validation Results} 
\label{tab:PRGCN}
\begin{center}
\begin{tabular}{|c|c|c|c|c|c|c|c|c|c|c|} 
\hline
Models & Citeseer & Cora & pubmed  \\
 \hline
PRGCN (1) & 70.3 & 82.0 & 77.7  \\
PRGCN (2) & 69.7 & 80.8 & 78.5  \\
PRGCN (3) & 68.5 & 80.6 & 78.5  \\
PRGCN (4) & 69.4 & 79.8 & 78.1  \\
PRGCN (5) & 68.7 & 80.4 & 75.1  \\
\hline
\end{tabular}
\end{center}
\end{table}

One issue with PRGCN is that the loss is actually decreasing as accuracy increases for these optimizations so the accuracy is clearly not correlated with loss after a point. This is a sign of overfitting, but I also believe that the transition matrix regularization is somewhat malformed or disconnected with the true properties of the structure of the class labels, which is a short-coming of assumption. The poor performance could also potentially be due to rounding errors and decoherence in the transition matrix at higher orders. \\

I do not think that this approach should be considered a successful regularization strategy. Performance is not improved and somewhat unpredictable. Furthermore, evaluating and multiplying a dense probabilistic matrix takes far longer than the graph laplacian in vanilla GCN. \\

\section{Conclusions}

This paper reproduced the results of \cite{tkipf} on the three standard Cora, Pubmed, and Citeseer datasets. Furthermore, this paper looked at 3 GCN variants in the hopes of improving performance. These variants did not outperform the vanilla GCN implementation, but they do elucidate some interesting insights into the graph training problem in the context of such a neural network architecture. \\

RGCN serves as a validation of the arguments in \cite{tkipf} that explicit regularization is not necessary. Regularization in this case does not generally improve performance, and furthermore the training time increases dramatically in terms of both the number of epochs and the time taken to evaluate the regularization. \\

PGCN shows us that using graph representations of different walk depths can certainly impact training performance. Furthermore, this impact varies with the structure of the dataset being trained. For example, shorted walk depths (k=1) help performance on simpler datasets such as the Citeseer dataset, while longer depths improve performance on larger datasets such as Pubmed. However, PGCN is generally outperformed by the vanilla GCN architecture. \\

The combination of higher order graph structure and regularization for PRGCN did not generally improve performance, The depth 1 approximation does seem to mimic the performance of vanilla GCN for the Cora dataset, but the lack of improvement at higher depths is concerning. Furthermore, the improvement seen at higher depths is not seen for Pubmed as it is in the standard RGCN variant. The author was hopeful that this more complicated model would be successful but it seems to be generally inferior and it takes a longer time to train. \\

This paper serves to show that there is an elegance in the simplicity of the vanilla GCN structure. This simplicity allows it to be robust, and to train quickly. The fact that it generally outperforms the variants here should be an indicator that the structure as it is presented in \cite{tkipf} has struck a chord in terms of scalability and performance. \\



\bibliography{Graph_Convolutional_Networks}
\bibliographystyle{icml2017}

\end{document}